# Quality Measures in Biometric Systems

Fernando Alonso-Fernandez[1], *Member, IEEE,* Julian Fierrez, *Member, IEEE*
Javier Ortega-Garcia, *Senior Member, IEEE*

*Abstract*— Biometric technology has been increasingly deployed in the last decade, offering greater security and convenience than traditional methods of personal recognition. But although the performance of biometric systems is heavily affected by the quality of biometric signals, prior work on quality evaluation is limited. Quality assessment is a critical issue in the security arena, especially in challenging scenarios (e.g. surveillance cameras, forensics, portable devices or remote access through Internet). Different questions regarding the factors influencing biometric quality and how to overcome them, or the incorporation of quality measures in the context of biometric systems have to be analysed first. In this paper, a review of the state-of-the-art in these matters is provided, giving an overall framework of the main factors related to the challenges associated with biometric quality.

*Index Terms*— Biometrics, security, quality assessment, sample quality.

## I. INTRODUCTION

The increasing interest on *biometrics* is related to the number of important applications where a correct assessment of identity is crucial. Biometrics refers to automatic recognition of an individual based on anatomical (e.g., fingerprint, face, iris, hand geometry) or behavioural characteristics (e.g., signature, gait, keystroke dynamics) [1]. Biometrics offers greater convenience and several advantages over traditional security methods based on something that you *know* (e.g. password, PIN) or something that you *have* (e.g. card, key). In biometric systems, users do not need to remember passwords or PINs, which can be forgotten, or carry cards or keys, which can be stolen.

Since the establishment of biometrics as a specific research area in late '90s, the biometric community has focused its efforts on the development of accurate recognition algorithms. Nowadays, biometric recognition is a mature technology, used in many government and civilian applications such as e-Passports, ID cards, or border control. Examples include the US-VISIT fingerprint system, the Privium iris system (Amsterdam Airport) or the SmartGate face system (Sydney Airport). But, during the last few years, the problem of quality measurement has emerged as an important concern in the biometric community after the poor performance observed on pathological samples [2]. It has been demonstrated by several studies and technology benchmarks that the performance of biometric systems is heavily affected by the quality of biometric signals e.g. see Figure 1. This operationally important step is nevertheless under-researched in comparison to the primary feature extraction or pattern recognition task. The performance degradation observed in less controlled situations is one of the main challenges facing biometric technologies [3]. The proliferation of portable hand-held devices with biometric acquisition capabilities or recognition at-a-distance and on-the-move are just two examples of non-ideal scenarios not yet sufficiently mature, which require robust recognition algorithms capable of handling a range of changing characteristics [1]. A quantitative example of the degradation observed in these scenarios can be seen in Figure 2. Another important example is forensics, in which intrinsic operational factors further degrade the recognition performance and are generally not replicated in controlled studies [4].

There are a number of factors that can affect the quality of biometric signals, and there are numerous roles of a quality measure in the context of biometric systems. Standardization bodies are also incorporating quality measures into existing data storage and exchange formats. This paper summarizes the state-of-the-art in the biometric quality problem, giving an overall framework of the different related factors.

## II. WHAT IS BIOMETRIC SAMPLE QUALITY?

It has not been until the last years that there is consensus about what biometric sample quality is. Broadly, a sample is of good quality if it is suitable for personal recognition. Recent standardization efforts (ISO/IEC 29794-1) have established three components of biometric sample quality, see Figure 3: *i) character* (inherent discriminative capability of the source), *ii) fidelity* (degree of similarity between a sample and its source, attributable to each step through which the sample is processed); and *iii) utility*, (impact of the individual biometric sample on the overall performance of a biometric system). The *character* of the sample source and the *fidelity* of the processed sample contribute to, or similarly detract from, the *utility* of the sample [3].

It is generally accepted that a quality metric should most importantly mirror the *utility* of the sample, so that samples assigned higher quality lead to better identification of individuals [3]. Thus, quality should be predictive of recognition performance. This statement, however, is largely subjective: not all recognition algorithms work equally (i.e. they are not based on the same features), and their performance is not affected by the same factors. For example, a face recognition algorithm "A" can be insensitive to illumination changes, whereas another algorithm "B" can be severely affected by changes in illumination. In this situation, a measure of illumination will be useful for predicting performance of "B", but not of "A". Therefore, an adequate quality measure will be largely dependent on the type of recognition algorithm considered. As the performance of different recognition algorithms may not be affected by the same signal quality factors, the efficacy of a quality estimation algorithm will be usually linked to a particular recognition algorithm, or thereof class.

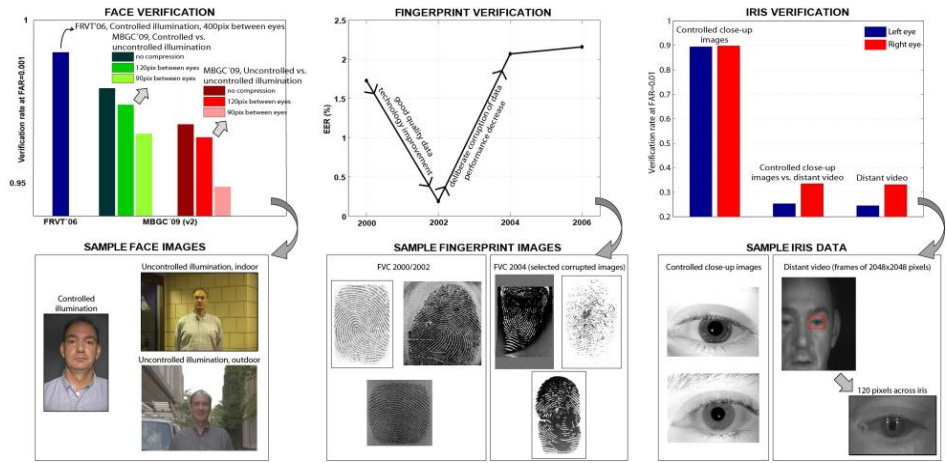

**Figure 1.** Effect of low quality data on the performance of recognition algorithms. Conditions progressively more difficult in nature result in a significant decrease in performance, in spite of the technology improvement between the different studies. Some sample images with varying quality are also shown in each modality. **Left**: best performing algorithm in face independent evaluations. FRVT stands for Face Recognition Vendor Technology, and MBGC for Multiple Biometric Grand Challenge. A decrease in performance is observed in the 2009 evaluation, when uncontrolled illumination conditions and severe image compression were introduced. More information at www.frvt.org and http://face.nist.gov/mbgc. **Middle**: best performing algorithm in the Fingerprint Verification Competitions (FVC). In 2000 and 2002, fingerprint data where acquired without any special restriction, resulting in an EER decrease of one order of magnitude. In the 2004 edition, samples were intentionally corrupted (e.g. by asking people to exaggeratedly rotate or press the finger against the sensor, or by artificially drying or moisturizing the skin with water or alcohol). More information at https://biolab.csr.unibo.it/fvcongoing. **Right**: results of the Video-based Automatic System for Iris Recognition (VASIR) implemented by the National Institute of Standards and Technology (NIST) on iris data from the MBGC. Performance on iris from distant video (unconstrained acquisition) is dramatically reduced with respect to classical close-up controlled acquisition. More information at http://www.nist.gov/itl/iad/ig/vasir.cfm

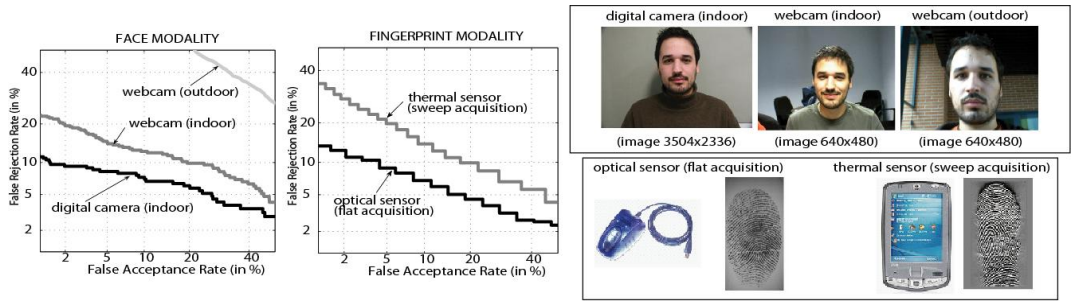

**Figure 2.** Performance degradation with portable devices. Face scores come from an LDA-based verifier using Fisher Linear Discriminant projection (face indoor) and an eigenface-based system with PCA analysis (face outdoor). Fingerprint scores come from the publicly available minutia-based matcher of the National Institute of Standards and Technology (NIST). Data is from the BioSecure Multimodal Database [5]. Face performance is degraded with the webcam, with further degradation in the more challenging outdoor environment (noisy ambience). As for the fingerprint modality, the sweep sensor results in worse performance with respect to the flat sensor. In flat sensors, acquisition is done by the touch method: the finger is simply placed on the scanner. In sweep sensors, the finger is swept vertically across a tiny strip with a height of only a few pixels. As the finger is swept, partial images are formed which are further combined to generate a full fingerprint image. This procedure allows to reduce the size and cost of the sensing element (facilitating its use in consumer products such as laptops, PDAs and mobile phones), but the reconstruction of a full image from the slices is prone to errors, especially in poor quality fingerprints and non-uniform sweep speed. (Figure extracted from Ortega-Garcia et al. [5])

### III. FACTORS INFLUENCING BIOMETRIC QUALITY

There are a number of factors affecting the quality of biometric signals, which are summarized in Table I. Unfortunately some of them are beyond control of system developers or operators. Therefore, assessing the quality of captured samples will allow appropriate corrective actions to take place. Following the framework of Kukula *et al.* [6] and contributions from other precedent works [7,8,9], a classification of quality factors based on their relationship with the different parts of the system is proposed [10]. Using this classification, four different classes can be distinguished:

**User-related factors**, which include *physical/ physiological* and *behavioural* factors. As they have to do entirely with the "user side", they are the most difficult to control. Some

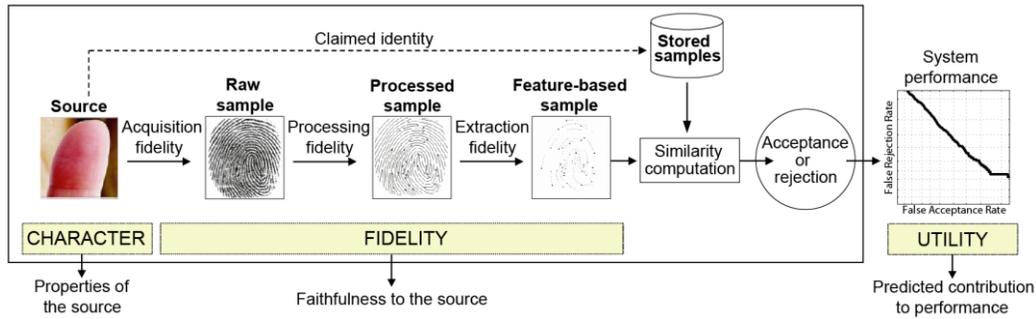

**Figure 3.** Definition of biometric quality from three different points of view: character, fidelity or utility.

*physical/physiological* factors inherent to the person (age, gender or race) do not produce degradation, but data variability that needs to be properly considered by the recognition algorithm (e.g. differences in speech between males and females). Diseases or injuries may alter face, finger, etc., even irreversibly, making them infeasible for recognition. Although, in some cases, the presence of such alterations can be precisely used to narrow a person's identity (e.g. amputation in gait recognition). On the other hand, coping with *behavioural* factors often implies *modifying* people's behaviour or habits, which is not always convenient, or is even impossible in some applications like forensics or surveillance cameras. Many behavioural factors can be alleviated by recapturing after taking corrective actions (e.g. "take off your hat/coat/ring/glasses" or "keep your eyes opened"), but this is not always possible. Depending on the application, corrective actions can result in people's reluctance to use the system. As can be seen in Table I, user-related factors have impact on the *character* of the biometric sample, that is, the quality attributable to inherent physical features. In this sense, the degree of control on these factors is low, as the inherent features of a person are difficult or impossible to modify. The remaining factors affect the *fidelity*, or in other words, the faithfulness between a biometric sample and its source, and their degree of control can be higher, as discussed next.

**Factors related to the user-sensor interaction**, which include *environmental* and *operational* factors. In principle, these are easier to control than user-related factors, although users still play a role in these. Users impact will depend on the level of control of the environment, the acquisition itself, and whether the acquisition physically takes place in controllable premises. In many applications, biometric data is acquired in less than ideal conditions, such as by surveillance cameras or portable hand-held devices. Other hot topic includes acquisition "at a distance" or "on the move" as a person walks by detection equipment, facilitating the ease of interaction with the system. But the unsatisfactory performance of biometrics technologies in these uncontrolled situations has limited their deployment, being one of the main challenges facing biometric technologies [1].

**Factors related to the acquisition sensor**. The sensor is in most cases the only physical point of interaction between the user and the biometric system. Its "fidelity" (see Section II) in reproducing the original biometric pattern is crucial for the accuracy of the recognition system. The diffusion of low cost sensors and portable devices (e.g. mobile cameras, webcams, telephones and PDAs with touch screen displays, etc.) is rapidly growing in the context of convergence and ubiquitous access to information and services, representing a new scenario for automatic biometric recognition systems. Unfortunately, these low cost and portable devices produce data which are very different from those obtained by dedicated (and more expensive) sensors, primarily due to a small input area, poor ergonomics or the fact that the user may be in movement. In this context, a measure of the reliability of the data and recognition process can provide additional improvement, by optimizing a structure lacking homogeneity, while ensuring system interoperability by integrating data of different nature [11].

**Factors related to the processing system**. Related to how a biometric sample is processed once it has been acquired, these are the factors, in principle, easiest to control. Storage or exchange speed constraints may impose the use of data compression techniques, e.g. smart cards. Also, governments, regulatory bodies, and international standards organizations often specify that biometric data must be kept in raw form, rather than in (or in addition to) post-processed templates that may depend on proprietary algorithms, with implications in data size. Hence, the effects of data compression on recognition performance become critical. The necessity for data compression, together with packet loss effects, also appears in recent applications of biometrics over mobile or Internet networks.

IV. ENSURING GOOD QUALITY OF BIOMETRIC SAMPLES

After analysing the usual factors affecting quality of biometric systems, this section reports some helpful guidelines for their control [7], which are summarized in Table II. Three points of action can be identified: *i*) the capture point, a critical point of action since it acts as the main interface between the user and the system, *ii*) the quality assessment algorithm itself, and *iii*) the system that performs the recognition process. If quality can be improved, either by capture point design or by system design, better performance can be realized. For those aspects of quality that cannot be designed-in, an ability to analyse the quality of a sample and initiate corrective actions is needed. This is useful, for example, in initiating the reacquisition from a user, selecting the best sample in real time, or selectively evoking different processing methods, and it is the key component in quality assurance management.

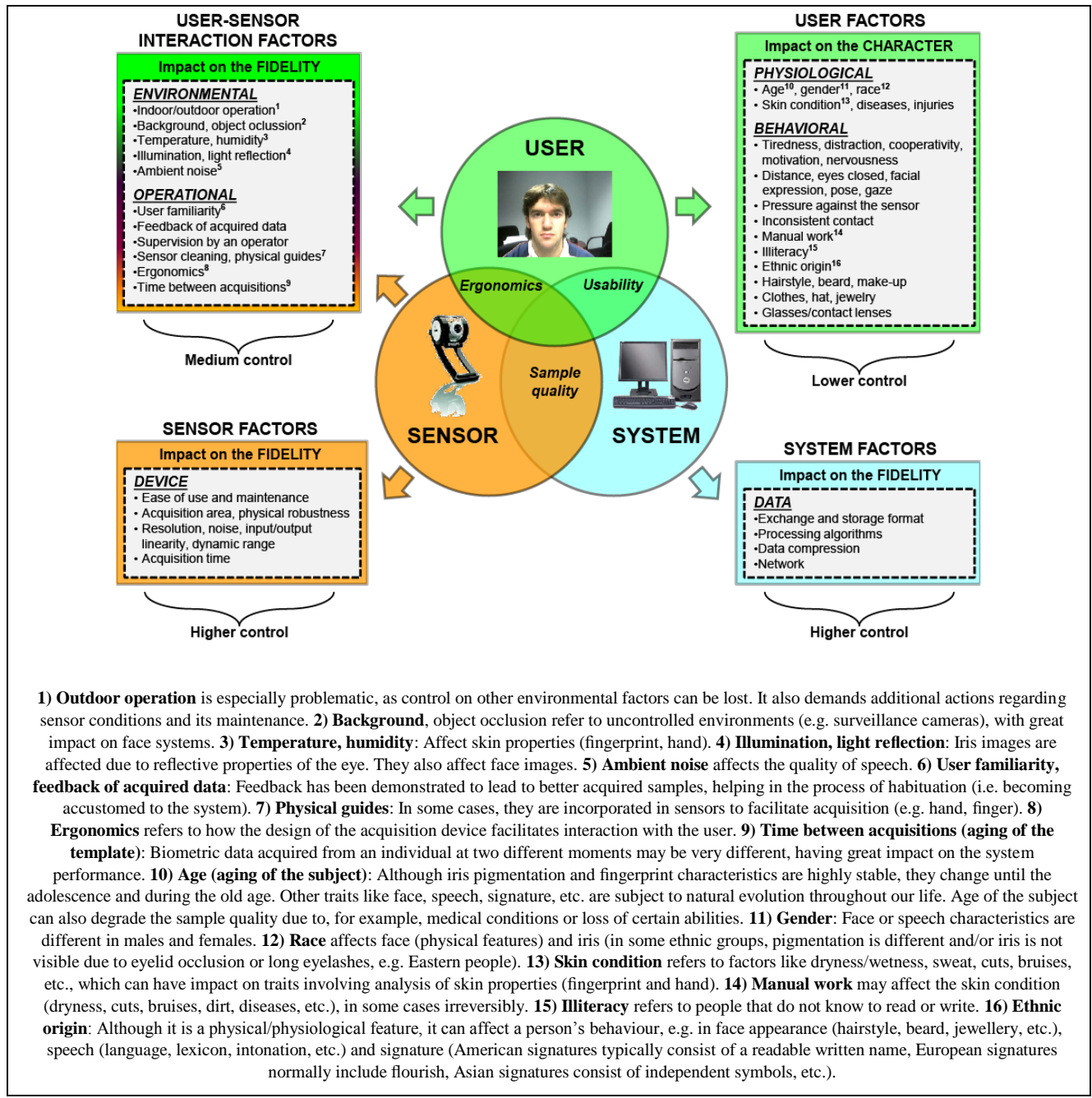

TABLE I
FACTORS AFFECTING THE QUALITY OF BIOMETRIC SIGNALS.

## V. QUALITY ASSESSMENT ALGORITHMS AND THEIR PERFORMANCE

Many quality assessment algorithms are found in the literature, focused on measuring different factors affecting the quality of biometric traits (see Figure 4). It is not the scope of this work to describe them in depth, so only a selection of key recent references is provided here (see references therein also). Quality assessment algorithms have been developed mainly for fingerprint images [14] and recently, for iris [15], voice [16], face [17] and signature signals [18]. In spite of the number of existing algorithms, almost all of them have been tested under limited and heterogeneous frameworks, mainly because it has not been until the last years when the biometric community has formalized the concept of sample quality and has developed evaluation methodologies. Two recent frameworks proposed for this purpose are briefly described here [3], [19].

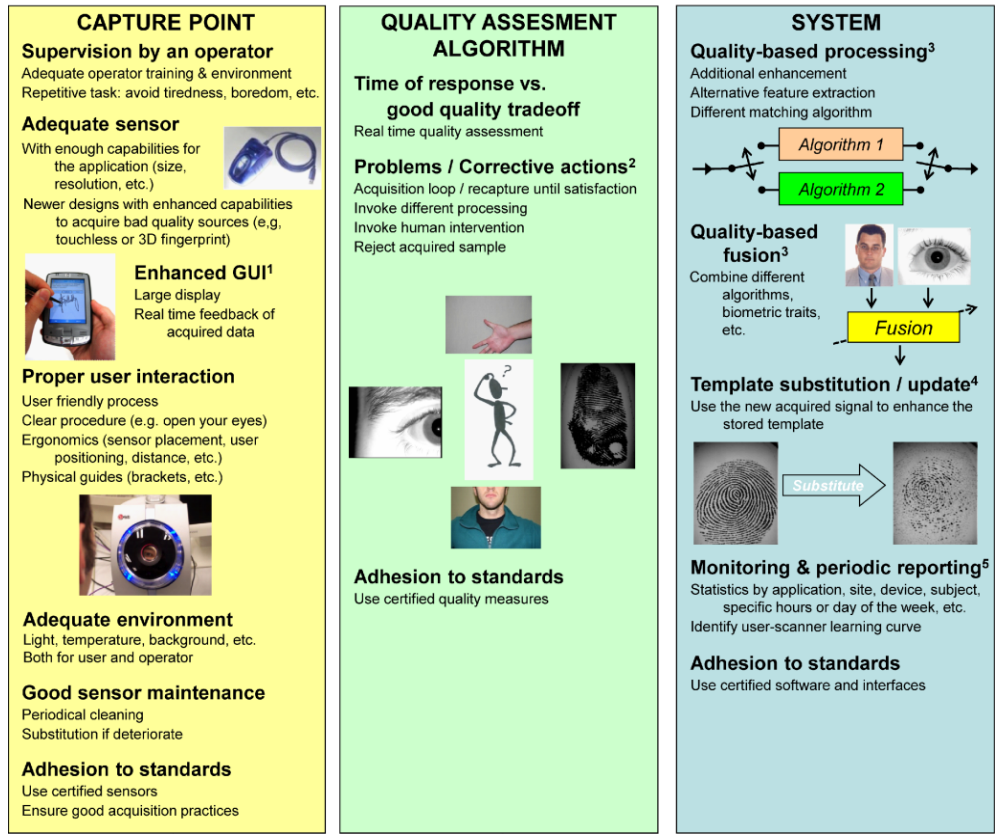

1) **Use of an adequate Graphical User Interface (GUI)**, with a large display providing real time feedback of acquired data, has demonstrated to help users to provide better signals over time and to habituate faster to the system [9]. 2) **Corrective actions** depend heavily on the application. For example, in some cases it is not possible to recapture a second sample (e.g. forensics), so the system has to deal with the "bad" sample at hand. Rejecting a sample implies invoking alternative recognition procedures (e.g. another biometric trait) or human intervention, resulting in increased costs and user inconvenience. 3) **Quality-based processing and fusion** means to invoke different algorithms and to combine them with different weighting depending on the quality of the signal at hand. See Section VII for further discussion. 4) **Template substitution/update**, an area still under-researched [12], allows coping with natural variations of biometric traits across time. Efficient strategies include storing multiple templates representative of the associated variability and updating/substituting them with new acquisitions. 5) **Monitoring and periodic reporting** [13] helps identify sudden problems (e.g. a damaged sensor) and find hidden systematic problems (e.g. specific sites or sensors working worse than others, hours when the quality of acquired signals is worse, etc.). Especially important is to identify user-scanner learning curves in order to avoid "first time user" syndrome, especially for elderly people or people who are not accustomed to interact with machines.

TABLE II
BIOMETRIC QUALITY ASSURANCE PROCESS.

As shown in Figure 3, biometric sample quality can be considered from the point of view of *character* (inherent properties of the source), *fidelity* (faithfulness of the biometric sample to the source), or *utility* (predicted contribution to performance). Youmaran and Adler [19] have developed a theoretical framework for measuring biometric sample *fidelity*. They relate biometric sample quality with the amount of identifiable information that the sample contains, and suggest that this amount decreases with a reduction in quality. They measure the amount of identifiable information for a person as the relative entropy, $D(p\|q)$, between the population feature distribution, $q$, and the person's feature distribution, $p$. Based on this, the information loss due to a degradation in sample quality can be measured as the relative change in the entropy.

On the other hand, most of the existing operational schemes for quality estimation of biometric signals are focused on the *utility* of the signal. Grother and Tabassi [3] have presented a framework for evaluating and comparing quality measures in terms of their capability of predicting the system performance. Broadly, they formalize the concept of sample quality as a scalar quantity that is related monotonically to the recognition performance of biometric matchers. Therefore, by partitioning the biometric data in different groups according to some quality criteria, the quality measure should give an ordered indication of performance between quality groups. Also, by rejecting low quality samples, error rates should decrease quickly with the fraction rejected. Some of the works referenced above in this Section have followed this framework in their experimental

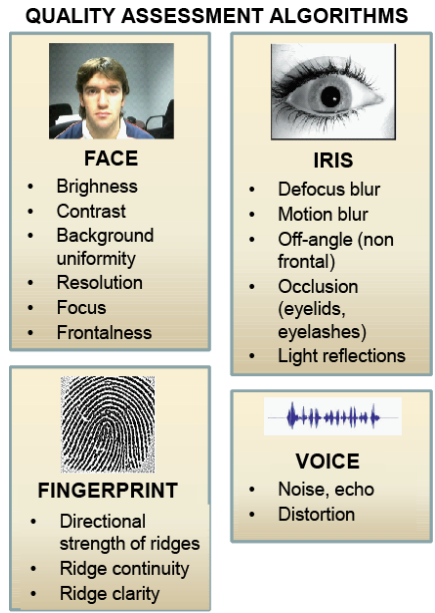

**Figure 4.** Common properties measured by biometric quality assessment algorithms. References to particular implementations are given in Section V.

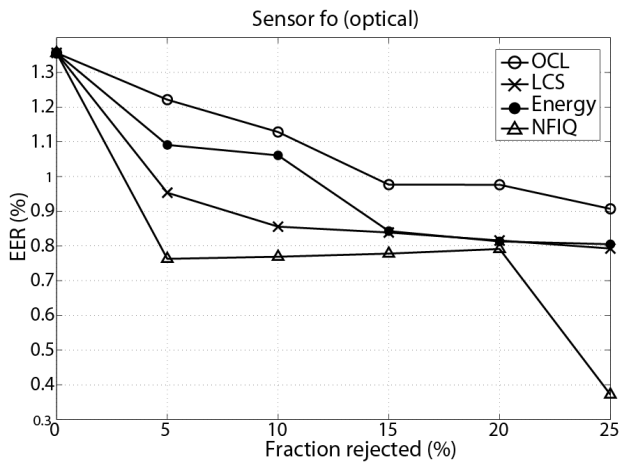

**Figure 5.** Evaluating the utility of four fingerprint quality measures. Results show the verification performance as samples with the lowest quality value are rejected. The similarity scores come from the publicly available minutia-based matcher released by the National Institute of Standards and Technology (NIST). Data is from the BioSec multimodal database [20]. Different performance improvement for the same fraction of rejected samples suggests different efficacy of each measure for the particular recognition algorithm and/or sensor evaluated. (Figure extracted from Alonso-Fernandez et al. [14])

studies. A graphical example evaluating the utility of fingerprint quality metrics can be seen in Figure 5. However, as mentioned before, the efficacy of a quality algorithm is usually tied to a particular recognition algorithm. This can be seen in the example of Figure 5, in which each quality metric results in different performance improvement for the same fraction of rejected low quality samples. It should be also noted that, although biometric matching involves at least two samples, they are not acquired at the same time. Reference samples are stored in the system database and are later compared with new samples provided during the operation of the system. Therefore, a quality algorithm should be able to work with individual samples, even though its ultimate intention is to improve recognition performance when matching two (or more) samples.

## VI. HUMAN VS. AUTOMATIC QUALITY ASSESSMENT

There is an established community of human experts in recognizing biometric signals for certain applications (e.g. signatures on checks or fingerprints in forensics) and the use of manual quality verification is included in the workflow of some biometric applications (e.g. immigration screening and passport generation). A common assumption here is that human assessment of biometric quality is an appropriate gold standard against which biometric sample quality measures should be measured [21]. Also, many authors make use of datasets with manually labelled quality measures to optimize and test their quality assessment algorithms.

To the best of our knowledge, the only study aimed to test the relevance of human evaluations of biometric sample quality is [21]. From this study, it is evident that human and computer processing are not always functionally comparable. For instance, if a human judges a face or iris image to be good because of its sharpness, but a recognition algorithm works in low frequencies, then the human statement of quality is inappropriate. The judgement of human inspectors can be improved by adequate training on the limitations of the recognition system, but this could be prohibitively expensive and time consuming. In addition, there are other implications in incorporating a human quality checker, such as tiredness, boredom or lack of motivation that a repetitive task like this may cause in the operator, as pointed out in Section IV. A comprehensive analysis of factors leading to errors related with human-assisted operation is given by Wertheim [22].

## VII. INCORPORATING QUALITY MEASURES IN BIOMETRIC SYSTEMS

The incorporation of quality measures in biometric systems is an active field of research, with many solutions proposed. Different uses of sample quality measures in the context of biometric systems have been identified throughout this paper. These are summarized in Table III [7], [8]. It should be noted that these roles are not mutually exclusive. Indeed, prevention of poor quality data requires a holistic, system-wide focus involving the whole operation of the biometric system [23]. It is not the scope of this paper to provide a comprehensive list of references. We refer the interested reader to the surveys contained in references [3], [10], [12], [13], [23].

## VIII. STANDARDIZING BIOMETRIC QUALITY

It should be noted that adhesion to standards for sensors, software, interfaces, etc. is recommended throughout the quality assurance process. With the use of standards, great flexibility and modularity is obtained, as well as fast technology interchange, sensor and system interoperability, and proper interaction with external security systems.

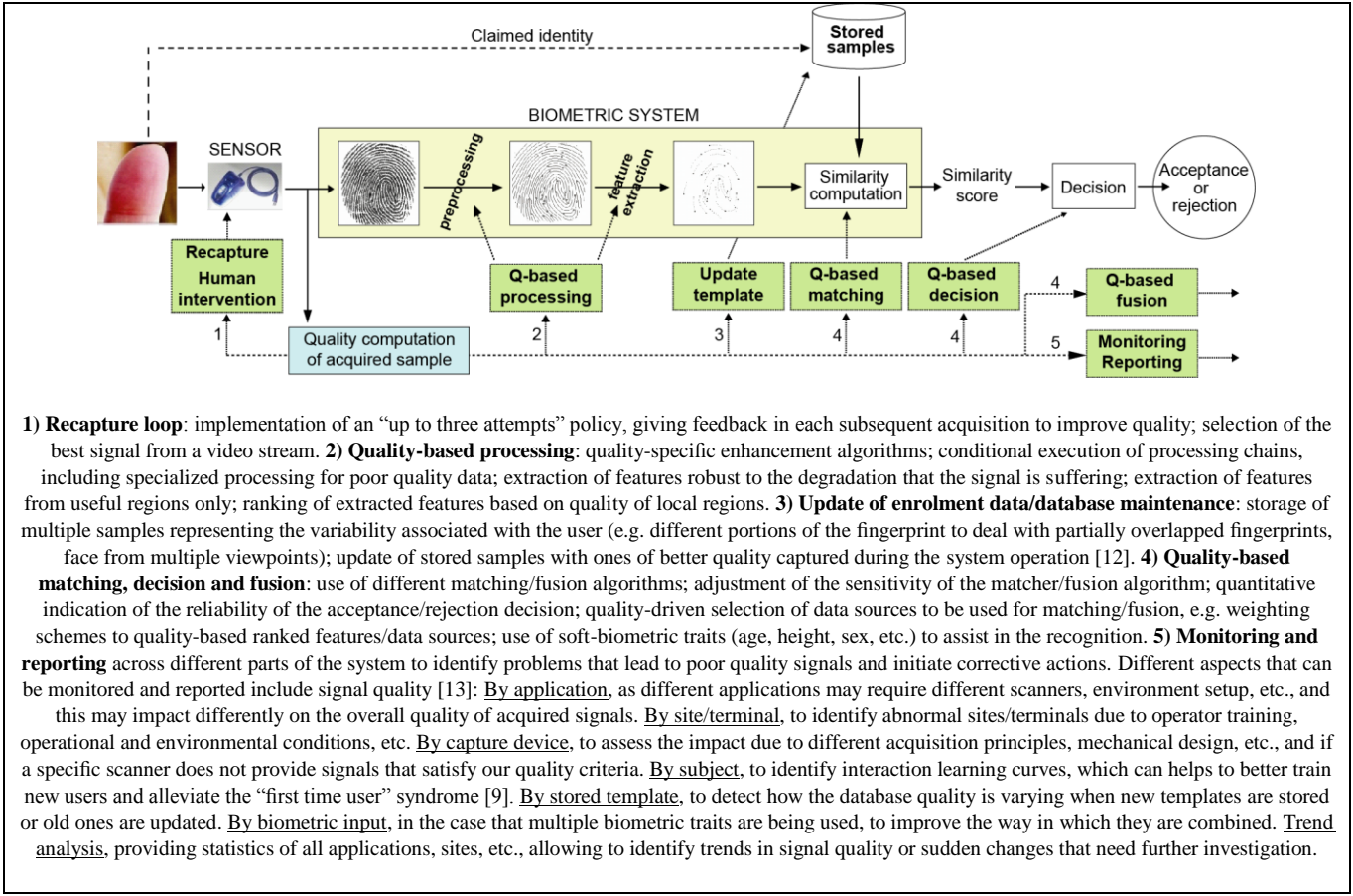

**1) Recapture loop**: implementation of an "up to three attempts" policy, giving feedback in each subsequent acquisition to improve quality; selection of the best signal from a video stream. **2) Quality-based processing**: quality-specific enhancement algorithms; conditional execution of processing chains, including specialized processing for poor quality data; extraction of features robust to the degradation that the signal is suffering; extraction of features from useful regions only; ranking of extracted features based on quality of local regions. **3) Update of enrolment data/database maintenance**: storage of multiple samples representing the variability associated with the user (e.g. different portions of the fingerprint to deal with partially overlapped fingerprints, face from multiple viewpoints); update of stored samples with ones of better quality captured during the system operation [12]. **4) Quality-based matching, decision and fusion**: use of different matching/fusion algorithms; adjustment of the sensitivity of the matcher/fusion algorithm; quantitative indication of the reliability of the acceptance/rejection decision; quality-driven selection of data sources to be used for matching/fusion, e.g. weighting schemes to quality-based ranked features/data sources; use of soft-biometric traits (age, height, sex, etc.) to assist in the recognition. **5) Monitoring and reporting** across different parts of the system to identify problems that lead to poor quality signals and initiate corrective actions. Different aspects that can be monitored and reported include signal quality [13]: By application, as different applications may require different scanners, environment setup, etc., and this may impact differently on the overall quality of acquired signals. By site/terminal, to identify abnormal sites/terminals due to operator training, operational and environmental conditions, etc. By capture device, to assess the impact due to different acquisition principles, mechanical design, etc., and if a specific scanner does not provide signals that satisfy our quality criteria. By subject, to identify interaction learning curves, which can helps to better train new users and alleviate the "first time user" syndrome [9]. By stored template, to detect how the database quality is varying when new templates are stored or old ones are updated. By biometric input, in the case that multiple biometric traits are being used, to improve the way in which they are combined. Trend analysis, providing statistics of all applications, sites, etc., allowing to identify trends in signal quality or sudden changes that need further investigation.

TABLE III
ROLES OF A SAMPLE QUALITY MEASURE IN THE CONTEXT OF BIOMETRIC SYSTEMS.

Standards compliance allows for replacement of parts of deployed systems with various technological options coming from open markets. As biometric technology is extensively deployed, a common situation is the exchange of information between several multi-vendor applications of different agencies, involving heterogeneous equipment, environments or locations [1]. In response to a need for interoperability, biometric standards have been developed to allow modular integration of products, also facilitating future upgrades to newer developments. Examples of interoperable scenarios are the use of e-Passports readable by different countries, or the exchange of lists of criminals among Security Forces. A list of standards organizations and other bodies working in biometric standards development is given in Table IV. Current efforts in developing biometric standards [24, 25] are focused on acquisition practices, sensor specifications, data formats and technical interfaces, as we plot in Figure 6 and Table V. In addition, although particularly aimed to the assistance of US federal agencies in its development and implementation of biometric programs, there is a "Registry of USG Recommended Biometric Standards" (www.biometrics.gov/Standards) with some high level guidance with respect to its implementation.

Concerning the specific incorporation of quality information, most of the standards define a quality score field aimed to incorporate quality measures. However, its content is not explicitly defined or is somewhat subjective due to the lack in consensus on i) how to provide universal quality measures interpretable by different algorithms or ii) what are the key factors that define quality in a given biometric trait. These problems are being addressed in the multipart standardization effort ISO/IEC 29794-1/4/5. A prominent approach within this standard is the Quality Algorithm vendor ID (QAID), which incorporates standardized data fields that uniquely identifies a quality algorithm, including its vendor, product code and version. QAID fields can be easily added to existing data interchange formats such as the Common Biometric Exchange Formats Framework (CBEFF), enabling a modular multi-vendor environment that accommodates samples scored by different quality algorithms in existing data interchange formats.

## IX. ISSUES AND CHALLENGES

This paper gives an overall framework of the main factors involved in the biometric quality measurement problem. The increasing development of biometrics in the last decade, related to the number of important applications where a correct assessment of identity is a crucial point, has not been followed by extensive research on biometric data quality [3]. A significant improvement in performance in less controlled

**International Standards Organizations**
- **ISO-JTC1/SC37**: International Organization for Standardization, Committee 1 on Information Technology, Subcommittee 37 for Biometrics (www.iso.org/iso/jtc1 sc37 home)
- **IEC**: International Electrotechnical Commission (www.iec.ch)

**National Standards Bodies (NSBS)**
- **ANSI**: American National Standards Institute (www.ansi.org)

**Standards-Developing Organizations (SDOS)**
- **INCITS M1**: InterNational Committee for Information Technology Standards, Technical Committee M1 on Biometrics (http://standards.incits.org/a/public/group/m1)
- **NIST-ITL**: American National Institute of Standards and Technology, Information Technology Laboratory (www.nist.gov/itl)
- **ICAO**: International Civil Aviation Organization (www.icao.int)

**Other NON-SDOS participating in standards development efforts**
- **BC**: Biometrics Consortium (www.biometrics.org)
- **IBG**: International Biometrics Group (www.ibgweb.com)
- **IBIA**: International Biometric Industry Association (www.ibia.org)
- **DoD-BIMA**: American Department of Defence, Biometrics Identity Management Agency (www.biometrics.dod.mil)
- **FBI-BCOE:** American Federal Bureau of Investigation, Biometric Centre of Excellence (www.biometriccoe.gov)

TABLE IV
STANDARDS ORGANIZATIONS AND OTHER BODIES WORKING IN BIOMETRIC STANDARDS DEVELOPMENT (ALL LINKS ACCESSED OCTOBER 2011).

situations is one of the main challenges facing biometric technologies [1]. Now that there is international consensus that a statement of a biometric sample's quality should be related to its recognition performance, efforts are going towards an harmonized and universal interpretation of quality measures by defining the key factors that need to be assessed in each biometric trait [25], and by setting good acquisition practices [7]. This will enable a competitive multi-vendor marketplace, allowing interoperability of multiple vendors' quality assessment algorithms.

A biometric system has to be resilient in processing data with heterogeneous quality yet providing good recognition performance. Although there are several corrective actions that can be performed to improve the quality of acquired signals [7], some factors fall out of our control or cannot be avoided. In this respect, specially challenging scenarios for biometrics are the ones based on portable devices, and/or remote access through Internet or acquisition at-a-distance. These are expected to work in an unsupervised environment, with no control on the ambient noise, on the user-sensor interaction process, or even on the sensor maintenance. Another very important field with inherent degraded conditions is forensics. Therefore, it is very important upon capture of biometric samples to assess their quality as well as having specific developments for poor quality signals [3].

Quality is intrinsically multi-dimensional, with factors of very different nature affecting it [6], [7], [8], [9]. A biometric system must adequately address this multifactor nature. There are a number of things that quality measures can do in

**BioApi** (Biometric Application Programming Interface), defines architecture and necessary interfaces to allow biometric applications to be integrated from modules of different vendors. Versions 1.0 and 1.1 were produced by the BioAPI Consortium, a group of over 120 companies and organizations with interest in biometrics. BioAPI 2.0 is specified in the ISO/IEC 19784-1 standard (published May 2006).

**CBEFF** (Common Biometric Exchange Formats Framework), supports exchange of biometrics information between different system components or systems. Developed from 1999 to 2000 by the CBEFF Development Team (NIST) and the BioAPI Consortium.

**FBI-WSQ** (FBI Wavelet Scalar Quantization) image compression algo-rithm for fingerprint images, developed to archive the large FBI fingerprint database. Developed by the FBI and the NIST.

**FBI-EBTS** (FBI Electronic Biometric Transmission Specification), **DoD-EBTS** (DoD Electronic Biometric Transmission Specification), **DHS-IDENT-IXM** (DHS Automated Biometric Identification System-Exchange Messages Specification) for exchange of biometric data with the FBI, DoD and DHS biometric applications, respectively. FBI-EBTS and DoD-EBTS are particular implementations of the ANSI/NIST ITL 1-2007 standard, customized to the needs of the FBI and the DoD. FBI-EBTS v9.2 released on May 2011. DoD-EBTS v2.0 released in March 2009. DHS-IDENT-IXM v5.0 released in November 2009.

**ANSI/NIST-ITL 1-2000** for exchange of biometric data between law enforcement and related criminal justice agencies, including fingerprint, facial, scar, mark, and tattoo data.

**ANSI/NIST-ITL 1-2007/2-2008** and **ISO/IEC-19794** multipart standard that specify a common format to exchange and store a variety of biometric data including face, fingerprint, palm print, face, iris voice and signature data.

**Annex to ISO/IEC-19794-5** with recommendations for face photo taking for E-passport and related applications, including indications about lighting and camera arrangement, and head positioning.

**ISO/IEC 29794-1/4/5 multi-part standard** to enable harmonized interpretation of quality scores from different vendors, algorithms and versions by setting the key factors that define quality in different biometric traits. It also addresses the interchange of biometric quality data via the multipart ISO/IEC 19794 Biometric Data Interchange Format Standard.

TABLE V
AVAILABLE BIOMETRIC STANDARDS (WITH RESPONSIBLE AGENCIES AND LATEST VERSION AVAILABLE).

the context of biometric systems to improve the overall performance, such as altering the sample processing/ comparison process, or weighting the results from different systems depending on the quality. Recent independent evaluations of commercial and research prototypes are also starting to include quality studies in their scenarios, as the BioSecure Multimodal Evaluation Campaign in 2007 (www.int-evry.fr/biometrics/BMEC2007) or the Noisy Iris Challenge Evaluation in 2009 (http://nice2.di.ubi.pt). Some research works have dealt with these matters, but much work is still to be done in this area.

ACKNOWLEDGEMENTS

Work of F.A.-F. at ATVS/Biometric Recognition Group has been supported by a Juan de la Cierva postdoctoral Fellowship from the Spanish MICINN. F. A.-F. also thanks

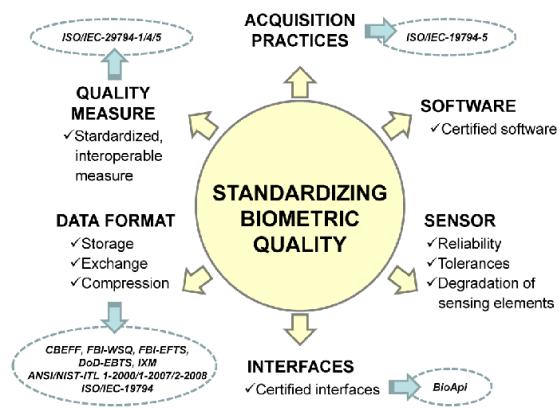

**Figure 6.** Use of standards in biometric systems to ensure good quality signals. See Table V for a more detailed description.

the Swedish Research Council (Vetenskapsrådet) and the European Commission for funding his postdoctoral research at Halmstad University. This work was also supported by projects Contexts (S2009/TIC-1485) from CAM, Bio-Challenge (TEC2009¬11186) from Spanish MICINN, TABULA RASA (FP7-ICT¬257289) and BBfor2 (FP7-ITN-238803) from EU, and Cátedra UAM-Telefónica. The authors would also like to thank the Spanish Dirección General de la Guardia Civil for their support to the work

**Fernando Alonso-Fernandez** received the M.S. degree in 2003 with Distinction and the Ph.D. degree "cum laude" in 2008, both in Electrical Engineering, from Universidad Politecnica de Madrid (UPM), Spain. Since 2004, he has been affiliated with the Biometric Recognition Group (ATVS), first working towards the Ph.D. degree, and later as postdoctoral researcher. He is currently a postdoctoral researcher at the Intelligent Systems Laboratory (IS-lab), Halmstad University, Sweden, under a postdoctoral fellowship of the Swedish Research Council (Vetenskapsrådet) and a Marie Curie fellowship of the European Commission. His research interests include signal and image processing, pattern recognition and biometrics. He has published several journal and conference papers and he has been actively involved in European projects focused on biometrics (e.g., Biosecure NoE, COST 2101). He has participated in the development of several systems for a number of biometric evaluations (e.g. SigComp 2009, LivDet 2009, BMEC 2007). Dr. Alonso-Fernandez has been invited researcher in several laboratories across Europe, and is the recipient of a number of distinctions for his research, including: best Ph.D. Thesis on Information and Communication Technologies applied to Banking in 2010 by the Spanish College of Telecommunication Engineers (COIT), and Doctorate Extraordinary Award in 2011 by Universidad Politecnica de Madrid to outstanding Ph.D. Thesis.

**Julian Fierrez-Aguilar** received the M.Sc. and the Ph.D. degrees in telecommunications engineering from Universidad Politecnica de Madrid, Madrid, Spain, in 2001 and 2006, respectively. Since 2002 he has been affiliated with the Biometric Recognition Group (ATVS), first at Universidad Politecnica de Madrid, and since 2004 at Universidad Autonoma de Madrid, where he is currently an Associate Professor. From 2007 to 2009 he was a visiting researcher at Michigan State University in USA under a Marie Curie fellowship. His research interests and areas of expertise include signal and image processing, pattern recognition, and biometrics, with emphasis on signature and fingerprint verification, multi-biometrics, biometric databases, and system security. Dr. Fierrez has been and is actively involved in European projects focused on biometrics, and is the recipient of a number of distinctions for his research, including: best Ph.D. thesis in computer vision and pattern recognition in 2005-2007 by the IAPR Spanish liaison (AERFAI), Motorola best student paper at ICB 2006, EBF European Biometric Industry Award 2006, and IBM best student paper at ICPR 2008.

**Javier Ortega-Garcia** received the M.Sc. degree in electrical engineering (Ingeniero de Telecomunicaci´on), in 1989; and the Ph.D. degree "cum laude" also in electrical engineering (Doctor Ingeniero de Telecomunicación), in 1996, both from Universidad Politécnica de Madrid, Spain. Dr. Ortega-Garcia is founder and co-director of ATVS research group. He is currently a Full Professor at the Escuela Politécnica Superior, Universidad Autónoma de Madrid, where he teaches Digital Signal Processing and Speech Processing courses. He also teaches a Ph.D. degree course in Biometric Signal Processing. His research interests are focused on biometrics signal processing: speaker recognition, face recognition, fingerprint recognition, on-line signature verification, data fusion and multimodality in biometrics. He has published over 150 international contributions, including book chapters, refereed journal and conference papers. He chaired "Odyssey-04, The Speaker Recognition Workshop", co-sponsored by IEEE. Since 2008 he is a Senior member of the IEEE.